\documentclass[11pt,a4paper]{article}
\usepackage[hyperref]{eacl2021}
\usepackage{times}
\usepackage{latexsym}
\usepackage{url}
\usepackage{pgffor}
\usepackage{graphicx}
\usepackage{xspace}
\usepackage{tabularx}
\usepackage{balance}
\usepackage{multirow}
\usepackage{microtype}
\usepackage{booktabs}
\usepackage[inline]{enumitem}
\usepackage{amssymb}

\newcommand{\corpusVis}{\textsc{EmoVis}\xspace}
\newcommand{\corpusHidden}{\textsc{EmoHide}\xspace}
\newcommand{\corpusAuto}{\textsc{AutoAppr}\xspace}

\newcommand{\emotionname}[1]{\textit{#1}}
\newcommand{\fear}{\emotionname{fear}\xspace}
\newcommand{\joy}{\emotionname{joy}\xspace}
\newcommand{\anger}{\emotionname{anger}\xspace}

\newcommand{\sadness}{\emotionname{sadness}\xspace}

\newcommand{\disgust}{\emotionname{disgust}\xspace}
\newcommand{\annoyance}{\emotionname{annoyance}\xspace}
\newcommand{\insecurity}{\emotionname{insecurity}\xspace}
\newcommand{\boredom}{\emotionname{boredom}\xspace}
\newcommand{\relaxation}{\emotionname{relaxation}\xspace}
\newcommand{\beauty}{\emotionname{beauty}\xspace}
\newcommand{\uneasiness}{\emotionname{uneasiness}\xspace}
\newcommand{\vitality}{\emotionname{vitality}\xspace}
\newcommand{\awe}{\emotionname{awe}\xspace}
\newcommand{\sublime}{\emotionname{sublime}\xspace}
\newcommand{\humor}{\emotionname{humor}\xspace}
\newcommand{\nostalgia}{\emotionname{nostalgia}\xspace}
\newcommand{\suspense}{\emotionname{suspense}\xspace}

\newcommand{\attention}{\textit{attention}\xspace}
\newcommand{\certainty}{\textit{certainty}\xspace}
\newcommand{\circumstance}{\textit{circumstance}\xspace}
\newcommand{\control}{\textit{control}\xspace}
\newcommand{\effort}{\textit{anticipated effort}\xspace}
\newcommand{\pleasantness}{\textit{pleasantness}\xspace}
\newcommand{\responsibility}{\textit{responsibility}\xspace}
\newcommand{\responsibilityC}{\textit{self-other responsibility/control}\xspace}
\newcommand{\situationalControl}{\textit{situational control}\xspace}

\newcommand{\F}{$\textrm{F}_1$\xspace}
\newcommand{\avg}{$\varnothing$}
\newcommand{\rt}[1]{\rotatebox{90}{#1}}

\newcommand{\taskTE}{T$\rightarrow$E\xspace}
\newcommand{\taskTA}{T$\rightarrow$A\xspace}
\newcommand{\taskAE}{A$\rightarrow$E\xspace}

\usepackage{trimclip}
\newcommand{\shortarrow}{\clipbox{5pt 0pt 0pt -1pt}{$\rightarrow$}}

\usepackage{scalefnt}
\usepackage{xcolor}
\usepackage{color,soul}

\setul{0.5ex}{0.3ex}
\setulcolor{red}

\aclfinalcopy
\def\aclpaperid{8}
\def\confidential{\textcolor{black}{WASSA 2021 Submission~\aclpaperid.  Confidential Review Copy.  DO NOT DISTRIBUTE.}}

\title{Emotion-Aware, Emotion-Agnostic, or Automatic:\\ Corpus
  Creation Strategies to Obtain\\ Cognitive Event Appraisal Annotations}

\author{Jan Hofmann, Enrica Troiano, \and Roman Klinger \\
  Institut f{\"u}r Maschinelle Sprachverarbeitung, University of Stuttgart, Germany \\
  \texttt{\{jan.hofmann,enrica.troiano,roman.klinger\}@ims.uni-stuttgart.de}\\
}

\date{}

\begin{document}
\maketitle
\begin{abstract}
  Appraisal theories explain how the cognitive evaluation of an event
  leads to a particular emotion. In contrast to theories of basic
  emotions or affect (valence/arousal), this theory has not received a
  lot of attention in natural language processing. Yet, in psychology
  it has been proven powerful: \newcite{Smith1985} showed that
  the appraisal dimensions \attention, \certainty, \effort,
  \pleasantness, \responsibility/\control and \situationalControl
  discriminate between (at least) 15 emotion classes.
  We study different annotation strategies for these dimensions,
  based on the event-focused enISEAR corpus
  \cite{Troiano2019}.  We analyze two manual annotation settings:
  (1)~showing the text to annotate while masking the experienced emotion
  label; (2)~revealing the emotion associated with the text. Setting 2
  enables the annotators to develop a more realistic intuition of the
  described event, while Setting 1 is a more standard annotation
  procedure, purely relying on text.
  We evaluate these strategies in two ways: by measuring
  inter-annotator agreement and by fine-tuning RoBERTa to predict
  appraisal variables. Our results show that knowledge of the emotion
  increases annotators' reliability. Further, we evaluate a purely
  automatic rule-based labeling strategy (inferring appraisal from
  annotated emotion classes). Training on automatically assigned
  labels leads to a competitive performance of our classifier, even
  when tested on manual annotations. This is an indicator that it
  might be possible to automatically create appraisal corpora for
  every domain for which emotion corpora already exist.
\end{abstract}

\section{Introduction}
Automatically detecting emotions in written texts consists of mapping
textual units, like documents, paragraphs, or sentences, to a
predefined set of emotions.  Common sets of classes used for this
purpose rely on psychological theories such as those proposed by
\newcite{Ekman1992} (\textit{anger}, \textit{disgust}, \textit{fear},
\textit{joy}, \textit{sadness}, \textit{surprise}) or
\newcite{Plutchik2001}. These theories are based on the assumption
that there is a restricted number of emotions that have prototypical
realizations. However, not all sets of emotions are
appropriate for every domain. For instance, \newcite{Dittrich2019}
argue that some of the basic emotions are too strong for measuring
how people feel when driving a car and, based on that,
\newcite{Cevher2019} resort to \joy, \annoyance (instead of \anger),
\insecurity (instead of \fear), \boredom, and \relaxation to classify
in-car utterances. \newcite{Haider2020} model the emotional perception
of poetry and opt for the categories \beauty/\joy, \sadness,
\uneasiness, \vitality, \awe/\sublime, \suspense, \humor, \nostalgia,
and \annoyance, following the definition of aesthetic emotions
\cite{Schindler2017,Menninghaus2019}. \newcite{Demszky2020} define a
taxonomy of emotions, reaching a high coverage while maintaining
inter-class relations.

An alternative to the use of categorical variables are the so-called
``dimensional'' approaches. The most popular of them models affective
experiences along the variables of dominance, valence, and arousal
\cite[VAD]{russell1977evidence}. \newcite{Barrett2006,Feldmanbarrett2017}
theorizes that emotions are interpretations of continuous affective
states experiencers find themselves in. Still, as \newcite{Smith1985}
note, not all emotions can be distinguished based on valence and
arousal. One might argue that predicting three continuous variables
instead of a richer set of categories is a simplification and can be
limiting for downstream applications of emotion analysis models.

\newcite{Smith1985} particularly argue that the VAD model does
not capture all relevant aspects of an emotion in the context of an
event. In a \textit{fight or flight} situation \cite{Cannon1929}, for
instance, the decision to take one of these two actions is mostly made
based on the effort that the emotion experiencer anticipates, but this
is not represented by VAD. Therefore, \newcite{Smith1985} propose a
dimensional approach with the appraisal variables of how pleasant an
event is (\pleasantness, likely to be associated with \joy, but
unlikely to appear with \disgust), how much effort an event can be
expected to cause (\effort, likely to be high when \anger or \fear is
experienced), how certain the experiencer is in a specific situation
(\certainty, low, e.g., in the context of \emotionname{hope} or
\emotionname{surprise}), how much attention is devoted to the event
(\attention, likely to be low, e.g., in the case of
\emotionname{boredom} or \emotionname{disgust}), how much
responsibility the experiencer of the emotion has for what has
happened (\responsibilityC, high for feeling \emotionname{guilt} or
\emotionname{pride}), and how much the experiencer feels to be
controlled by the situation (\situationalControl, high in \anger).

As the cognitive appraisal is a fundamental subcomponent of emotions,
we deem that appraisal dimensions are useful to perform emotion 
recognition, and that even the prediction of appraisals themselves 
can contribute to computational approaches to affective states. 
\begin{table}
  \centering\small
  \begin{tabularx}{1.0\linewidth}{lp{13mm}X}
    \toprule
    Emotion & Appraisals & Text \\
    \cmidrule(r){1-1}\cmidrule(rl){2-2}\cmidrule(l){3-3}
    Joy & Attention, Certainty, Pleasant, Sit.\ Ctrl. & I felt \ldots\ when I knew that I was going back to Florida a year earlier than I thought I would. \\
    Disgust & Attention, Certainty, Effort,\par Sit.\ Ctrl. & I felt \ldots\ when my kitten was sick and I had to clean it up.\\
    Fear & Attention, Effort,\par Sit.\ Ctrl. & I felt \ldots\ when I was having a hard attach.\\
    Guilt &Attention, Certainty, Effort, Respons., Control &  I felt \ldots\ when I went on holiday and left our cat behind.\\
    Sadness &Attention, Certainty,\par Sit.\ Ctrl.& I felt \ldots\ when I found out one of my favourite shops had shut down.\\
    \bottomrule
  \end{tabularx}
  \caption{Examples from the corpus of \newcite{Hofmann2020}.
  }
  \label{tab:firstexample}
\end{table}
These appraisal dimensions have only recently found application in
automatic emotion analysis in text: \newcite{Hofmann2020} re-annotated
a corpus of 1001 English emotional event descriptions
\cite{Troiano2019} for which the experienced emotion has been
disclosed by the author of the description
(Table~\ref{tab:firstexample} shows examples from their corpus). Their
annotation is designed as a preliminary step for inferring discrete emotion
categories. In contrast, we argue \emph{that the prediction of appraisal
  dimensions is in itself valuable}. This intuition has an impact on our
annotation strategy. While \newcite{Hofmann2020} did not show any
emotion label to the annotators, thus avoiding information
leaks, we hypothesize that knowing such emotion 
helps understanding how the described events were originally appraised by their
experiencers: at times, properly annotating appraisals 
as a third party might be unfeasible without having prior access to emotions.

We test this by comparing three annotation procedures: (1)~we give the
annotator access only to the text but not to its emotion label; (2)~we give the
annotator access to the text and the emotion, and evaluate if this
additional information has an impact on annotation reliability and
performance of a pretrained transformer-based classifier fine-tuned on
these data; and (3)~we automatically infer the appraisal dimensions
from existing emotion annotations, investigating the hypothesis that
manual annotation might not be necessary.

Our main contributions are that we show that (a)~appraisal annotation
is more reliable when annotators have access to the emotion label of
the original experiencer, hence, \emph{the event description itself
  does not carry sufficient information for annotation}. That also
means that annotating appraisals for corpora in which the original
emotion is not available might be particularly challenging.
(b)~Automatic, rule-based annotation of appraisals that leverages
emotion labels is a viable alternative to human annotation, and
therefore, appraisal corpora can automatically be created for domains
for which emotion corpora are already available.

Our classifier further constitutes a novel state of the art for
appraisal prediction on the data by \newcite{Hofmann2020}. The data is
available at
\url{https://www.ims.uni-stuttgart.de/data/appraisalemotion}.

\section{Related Work}
\subsection{Resources for Emotion Analysis}
There is a wealth of literature in psychology surrounding emotions,
specifically regarding the way they are elicited, their universal
validity, their number and stereotypical expressions, and their
function \cite{Scherer2000,Gendron2009}. The two prominent traditions
which have dominated the field of emotion classification in natural
language processing are discrete and dimensional models
\cite{Kim2019b}.

Next to the creation of lexicons for emotion analysis
\cite[i.a.]{Pennebaker2001,Strapparava2004,Mohammad2013,Mohammad2018b},
the annotation of text corpora received substantial attention
\cite{Bostan2018}. They vary across emotion categories and domains,
with discrete classes being dominating -- some exceptions focused on
valence and arousal annotations are \newcite{Buechel2017},
\newcite{Preotiuc2016}, and \newcite{Yu2016}.  For instance, the ISEAR
study by \newcite{Scherer1994} led to self-reports of emotionally
connotated events. Its creators aimed at understanding what aspects of
emotions are universal and which are relative to culture. It was built
by asking students to recall an emotion-inducing event and to describe
it.

Other efforts focused more on creating corpora specifically for
emotion analysis in NLP. \newcite{Troiano2019} built enISEAR and
deISEAR, whose 1001 event-descriptions were collected via
crowd-sourcing, with a questionnaire inspired by ISEAR, both in
English and in German. TEC \cite{Mohammad2012}, another popular
resource, is bigger in size ($\approx$21k instances), contains tweets
and was automatically annotated with hashtags. The Blogs corpus by
\newcite{Aman2007} has sentence-level annotations for 5205
texts, annotated by multiple raters.  While ISEAR, enISEAR and
deISEAR are focused on describing specific emotion-inducing events,
the Blogs corpus and TEC are more general.

This is also the set of corpora that we use in our study 
(a more comprehensive resource overview was made available by
\newcite{Hakak2017} and \newcite{Bostan2018}).

\subsection{Appraisal Theories}
A richer perspective on emotions and their experience than affect
models or fundamental emotion sets is provided by appraisal models
\cite{Scherer2009}, which did not receive a lot of attention from the
NLP community so far.  Appraisals are immediate evaluations of
situations which guide the emotion felt by the experiencer
\cite{Scherer2009dynamic}. More precisely, an emotion is a
synchronized change in five organismic subsystems (i.e., cognitive,
peripheral efference, motivational, motor expression and subjective
feeling) in response to the evaluation of a stimulus event important
to an individual. Emotion states can be distinguished on the basis of
their accompanying appraisals. For instance, fear emerges when an
event is appraised as unforeseen and disagreeable, a frightening event
is one appraised as an unforeseen, unpleasant, and contrary to one's
goal \cite{mortillaro2012advocating}.  The cognitive part of the
emotion is the one guiding the evaluation of the stimulus along
different dimensions. According to \newcite{scherer2001}, they are
relevance (i.e., the pleasantness of the event, and its relevance for
one's goals), implication (i.e., its potential consequences), coping
potential (i.e., one's ability to adjust to or control the situation)
and normative significance (i.e. its congruity to one's values and
beliefs). On a similar vein, \newcite{Smith1985} argue that six
cognitive appraisal dimensions can differentiate emotional
experiences, as there is a relationship between the way situations are
appraised along such dimensions and the experienced emotion. They are
\pleasantness, \effort, \certainty, \attention,
\responsibility/\control and \situationalControl.  We use their model
to explore appraisals in text.

\section{Experimental Setting}
\subsection{Annotation Guidelines}
We adhere to the annotation guidelines and the
appraisal dimensions of \newcite{Hofmann2020}, splitting the
original \situationalControl from \newcite{Smith1985} into \control and \circumstance. 
Our judges take binary decisions with
respect to seven appraisal dimensions. We ask them the following
questions:

``Most probably, at the time when the event happened, the writer\ldots
\begin{itemize*}[topsep=0pt,itemsep=0pt,parsep=0pt,partopsep=0pt,label={;  }]
\item [] \ldots wanted to devote further attention to the event (\textit{Attention})
\item \ldots was certain about what was happening (\textit{Certainty})
\item \ldots had to expend mental or physical effort to deal with the
  situation (\textit{Effort})
\item \ldots found that the event was pleasant (\textit{Pleasantness})
\item \ldots was responsible for the situation (\textit{Responsibility})
\item \ldots found that he/she was in control of the situation (\textit{Control})
\item \ldots found that the event could not have been changed or
  influenced by anyone (\textit{Circumstance}).''
\end{itemize*}

\subsection{Experiment 1: Manual Annotation}
To annotate the appraisal dimensions, judges need to make
assumptions about the experienced situation. We believe this is possible,
at times, purely from the textual description that needs to be judged.
Other times, knowing which emotion a person
developed might be necessary to understand how the
overall experience was originally appraised.

To analyze this assumption and measure the importance of emotion
labels to reliably
assign appraisal dimensions, we build our experiment on top of the
English enISEAR corpus by \newcite{Troiano2019}. Its authors
asked workers on a crowdsourcing platform to complete sentences like
``I felt [emotion name], when/that/because\ldots'', where [emotion
name] is replaced by a concrete emotion. In a later annotation round,
other annotators had to infer the emotion of the text, and for this
reason the creators of the corpus replaced emotion words with
``\ldots''.  The resulting 1001 instances of enISEAR are labeled by
the experiencers of the emotion themselves and have masked
emotion words in the text.

We use these data to perform two annotation experiments on 210
instances, randomly sampled from enISEAR and stratified by
emotion. Two annotators judge all of these instances in two different
settings. Setting 1, \corpusHidden, replicates the study by
\newcite{Hofmann2020}: the emotion label is not available to the
annotator. In Setting 2, \corpusVis, the emotion is presented along
with the text. The two rounds of annotations (first \corpusHidden,
later \corpusVis which makes the emotion available) were
distantiated by six months, to avoid a bias by recalling the previous
round. We evaluate the reliability of the annotation via
inter-annotator agreement with \citeauthor{Cohen1960}'s $\kappa$
\shortcite{Cohen1960}, under the hypothesis that having knowledge of
the emotion leads to more reliable human annotations.

\textbf{Computational Modelling.} One of the annotators annotated the
full 1001 instances twice, that is, for the \corpusHidden and the
\corpusVis approaches as a basis to evaluate how well the realization
of the appraisal concepts in the corpus can be modelled
automatically. As we expect the annotations \corpusVis to be more
reliable, we also expect the model to perform better.\footnote{We
  acknowledge that model performance on an annotated corpus can only
  to some degree be used to assess data quality. However, in
  combination with our inter-annotator agreement assessments, it
  serves as an indicator of the amount of noise.}

We use RoBERTa \cite{Liu2019} with the abstraction layer for
tensorflow as provided by ktrain \cite{Maiya2020}, and choose the
number of epochs to be 5, based on the appraisal prediction and
emotion classification tasks in the data by \newcite{Hofmann2020}
(which we annotate in this paper). Only minor differences in performance
can be seen between epochs 4--7. We keep this number of epochs fixed across all
experiments and all other parameters at their default. The batch size
is 5. More concretely, we opt for a 3$\times$10-fold cross-validation
setting, use the RoBERTa-base model in all experiments except for
those on the German deISEAR (in the next Experiment 2), where we use
XLM-R \cite{Conneau2020}.

\subsection{Experiment 2: Automatic Annotation}
\begin{table}[t]
  \newcommand{\gz}{\textcolor{black!40!white}{0}}
  \centering\small
  \setlength{\tabcolsep}{10pt}
  \begin{tabular}{lcccccc}
    \toprule
    Emotion & \rt{Attention} & \rt{Certainty} & \rt{Effort} & \rt{Pleasant} & \rt{Resp/Contr.} & \rt{Sit.\ Control} \\
    \cmidrule(r){1-1}\cmidrule(lr){2-2}\cmidrule(lr){3-3}\cmidrule(lr){4-4}\cmidrule(lr){5-5}
    \cmidrule(lr){6-6}\cmidrule(l){7-7}
    Anger    &   1 &   1 &   1 & \gz & \gz & \gz \\
    Disgust  & \gz &   1 &   1 & \gz & \gz & \gz \\
    Fear     &   1 & \gz &   1 & \gz & \gz &   1 \\
    Guilt    & \gz &   1 &   1 & \gz &   1 & \gz \\
    Joy      &   1 &   1 & \gz &   1 &   1 & \gz \\
    Sadness  & \gz &   1 & \gz & \gz & \gz &   1 \\
    Shame    & \gz & \gz &   1 & \gz &   1 & \gz \\
    Surprise &   1 & \gz & \gz &   1 & \gz &   1 \\
    \toprule
  \end{tabular}
  \caption{Discretized associations between appraisal dimensions and
    emotion categories, following \protect\newcite{Smith1985}, as we use them for automatic annotation in Exp.~2.}
  \label{tab:automatic_annotation}
\end{table}
\begin{table*}[t]
  \centering\small
  \setlength{\tabcolsep}{9pt}
  \begin{tabular}{lcccc cccccc c}
    \toprule
    &\multicolumn{3}{c}{Inter Annotator Agreement} & \multicolumn{7}{c}{RoBERTa Modelling} \\
    \cmidrule(lr){2-4}\cmidrule(l){5-11}
    & \multicolumn{1}{c}{\corpusVis} & \multicolumn{1}{c}{\corpusHidden} && \multicolumn{3}{c}{\corpusVis} & \multicolumn{3}{c}{\corpusHidden} \\
    \cmidrule(lr){2-2}\cmidrule(lr){3-3}\cmidrule(lr){5-7}\cmidrule(lr){8-10}
    Appraisal & $\kappa$ & $\kappa$& $\Delta$ & P & R& \F & P & R & \F & $\Delta$\F\\
    \cmidrule(r){1-1}\cmidrule(lr){2-2}\cmidrule(lr){3-3}\cmidrule(lr){4-4}\cmidrule(lr){5-7}\cmidrule(rl){8-10}\cmidrule(l){11-11}
    Attentional Activity & .55 & .30 & $+$.25 &  .79 &.84 &.82  &  .84 &.88 &.86  & $-$.04 \\
    Certainty            & .71 & .43 & $+$.28 &  .94 &.97 &.96  &  .81 &.93 &.87  & $+$.09 \\
    Anticipated Effort   & .44 & .38 & $+$.06 &  .77 &.83 &.80  &  .66 &.58 &.62  & $+$.18 \\
    Pleasantness         & .93 & .87 & $+$.06 &  .92 &.94 &.93  &  .91 &.92 &.92  & $+$.01 \\
    Responsibility       & .80 & .64 & $+$.16 &  .85 &.79 &.82  &  .83 &.81 &.82  & $\pm$.00  \\
    Control              & .66 & .71 & $-$.05 &  .64 &.49 &.56  &  .74 &.68 &.71  & $-$.15 \\
    Circumstance         & .65 & .54 & $+$.11 &  .80 &.72 &.76  &  .76 &.74 &.75  & $+$.01 \\
    \cmidrule(r){1-1}\cmidrule(lr){2-2}\cmidrule(lr){3-3}\cmidrule(lr){4-4}\cmidrule(lr){5-7}\cmidrule(rl){8-10}\cmidrule(l){11-11}
    Macro \avg           & .68 & .55 & $+$.13 &  .82 &.80 &.80  &  .79 &.79 &.79  & $+$.01 \\
     Micro \avg          &     &     &        &  .84 &.85 &.84  &  .80 &.81 &.81  & $+$.03 \\
    \toprule
  \end{tabular}
  \caption{Experiment 1: Cohen's $\kappa$ between annotators on
    \corpusVis and \corpusHidden and modelling experiments. The model is separately trained and tested on \corpusVis and \corpusHidden.}
  \label{tab:exp1}
\end{table*}
As \newcite{Smith1985} showed, appraisal dimensions are sufficient to
discriminate emotion categories: this is knowledge which we can make
use of, and we can leverage their findings to automatically assign
discrete appraisal labels to enISEAR (see
Table~\ref{tab:automatic_annotation}) in a rule-based manner. For
comparability to the manual annotation setup, we opt for discrete
labels which we infer from the continuous principle component analysis
values from the original paper.

The question to be answered is if this rule-based annotation actually
represents the same concepts as the manual annotation. To answer this,
we compare the automatic annotation (\corpusAuto) with both
annotations that have been performed manually (\corpusHidden,
\corpusVis).  Further, we train a model to predict these automatic
annotations and evaluate on the manually annotated labels.

Since the automatic method relies on emotion labels, we expect its
annotations to be more similar to \corpusVis, where the annotators
also have access to this information. For the same reason, we also
assume that the model trained on automatically annotated labels
performs better on \corpusVis than on \corpusHidden.  Finding that
models trained on labels assigned in such rule-based manner show
comparable performance to manual annotations (when tested on manual
annotations) would suggest that the latter might not be necessary to
obtain appraisal prediction models.

In this automatic setup, we merge \responsibility and \control. While
they are divided in the manually annotated corpora, this separation is
not available in the results by \newcite{Smith1985}. This affects the
comparability of the averages of performance measures between Exp.~1
and~2.

\begin{figure*}
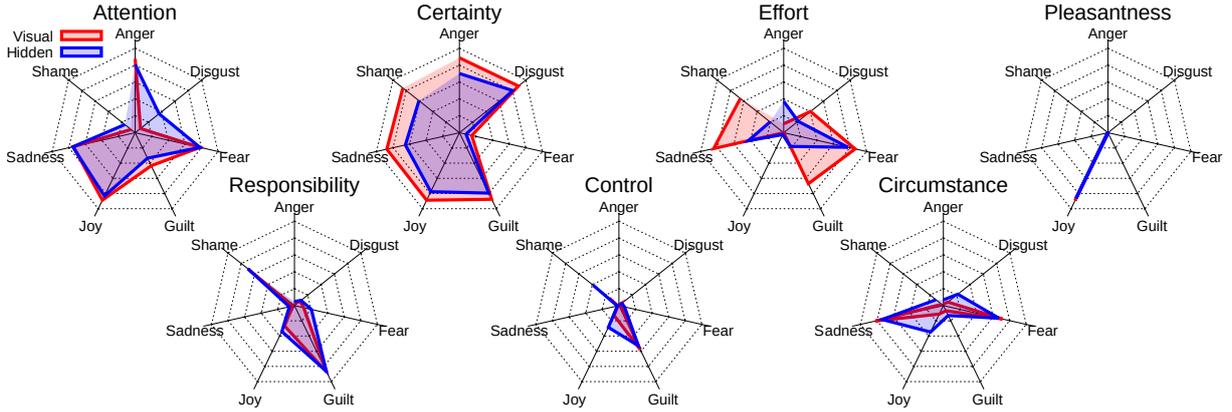

  \centering
  \foreach \n in {1,...,4}{%
    \includegraphics[width=0.20\linewidth,page=\n]{comparative-crop}\hfill
  }\\[-8mm]
  \mbox{}\hspace{20mm}
  \foreach \n in {5,...,7}{%
    \includegraphics[width=0.20\linewidth,page=\n]{comparative-crop}\hfill
  }
  \hspace{10mm}\mbox{}
  \caption{Frequency distribution of appraisals across emotions for \corpusVis (Visual)
    and \corpusHidden (Hidden).}
  \label{fig:dist}
\end{figure*}

Further, under the assumption that automatic annotation shows competitive
results on the manually annotated corpus enISEAR, we extend this
analysis to other resources for corpus generalization.
In addition to enISEAR, we use the original ISEAR dataset
\cite{Scherer1994}, the German event corpus deISEAR
\cite{Troiano2019} and, as resources without focus on
events, the Twitter Emotion Corpus (TEC) \cite{Mohammad2012} and the
Blogs corpus \cite{Aman2007}.
Since these corpora are not manually annotated for appraisals, we only
evaluate on automatic appraisal annotations.

\begin{table*}[t]
  \centering\small
  \setlength{\tabcolsep}{7.5pt}
  \begin{tabular}{lrrrr rrr rrr rrr}
    \toprule
    &\multicolumn{4}{c}{Agreement to \corpusAuto, $\kappa$}
    &\multicolumn{9}{c}{Modelling: RoBERTa trained on \corpusAuto} \\
    \cmidrule(r){2-5}\cmidrule(l){6-14}
    & \multicolumn{2}{c}{\corpusVis} & \multicolumn{2}{c}{\corpusHidden}   & \multicolumn{3}{c}{\corpusVis} & \multicolumn{3}{c}{\corpusHidden} & \multicolumn{3}{c}{\corpusAuto}\\
    \cmidrule(r){2-3}\cmidrule(lr){4-5}\cmidrule(lr){6-8}\cmidrule(rl){9-11}\cmidrule(l){12-14}
    Appraisal & A1 & A2 & A1 & A2 & P & R& \F & P & R & \F & P & R & \F\\
    \cmidrule(r){1-1}\cmidrule(r){2-3}\cmidrule(lr){4-5}\cmidrule(lr){6-8}\cmidrule(rl){9-11}\cmidrule(l){12-14}
    Attentional Activity   & .47 & .63 & .33 & .48 & .85 &.59 &.70 & .88 &.59 &.71 & .86 &.85 &.85\\
    Certainty              & .35 & .53 & .50 & .51 & .97 &.81 &.88 & .81 &.83 &.82 & .88 &.90 &.89\\
    Anticipated Effort     & .31 & .08 & .19 & .16 & .61 &.77 &.68 & .39 &.79 &.52 & .95 &.96 &.96\\
    Pleasantness           & .93 & 1.00& .91 & .96 & .90 &.93 &.92 & .91 &.95 &.93 & .96 &.94 &.95\\
    Responsibility/Control & .39 & .63 & .39 & .59 & .66 &.81 &.72 & .74 &.78 &.76 & .91 &.89 &.90\\
    Situational Control    & .44 & .64 & .34 & .59 & .77 &.74 &.75 & .78 &.59 &.67 & .84 &.83 &.84\\
    \cmidrule(r){1-1}\cmidrule(r){2-3}\cmidrule(lr){4-5}\cmidrule(lr){6-8}\cmidrule(rl){9-11}\cmidrule(l){12-14}
      Macro \avg           & .48 & .58 & .44 & .54 & .79 &.77 &.78 & .75 &.75 &.74 & .90 &.89 &.90\\
    Micro \avg             &     &     &     &     & .78 &.75 &.77 & .70 &.73 &.72 & .90 &.90 &.90\\
    \toprule
  \end{tabular}
  \caption{Experiment 2 Main Results: Cohen's $\kappa$ between annotators of
    \corpusHidden/\corpusVis and \corpusAuto, on the subset of
    210 instances from enISEAR. The
    classifier is trained
    on the full set of 1001 instances annotated automatically
    (\corpusAuto) and evaluated on all other annotations
    (cross-validation splits remain the same).}
  \label{tab:exp2}
\end{table*}

\section{Results}
\subsection{Experiment 1: Manual Annotation}
\paragraph{Inter-Annotator Agreement.}
In Experiment 1, we compare the reliability of the annotation with and
without access to the emotion label. We show the inter-annotator
agreement results in Table~\ref{tab:exp1}.
As we hypothesized, the agreement on \corpusVis is clearly higher than
on \corpusHidden, with .68 in comparison to .55 $\kappa$. The highest
agreement increase is observed for \attention ($+$.25) and \certainty
($+$.28), followed by \responsibility ($+$.16). The only decrease in
agreement, for \control, is comparably small ($-$.05).

Figure~\ref{fig:dist} (and Table~\ref{tab:manual_annotation_counts} in
the Appendix) shows the distribution of emotions for the
different appraisal dimensions: for most dimensions, the annotation
becomes more clearly connected to emotions with its availability, with
\certainty and \effort being exceptions: here, the number of instances
of a set of emotion classes partially increases. This confirms that
knowledge of the emotion ``denoises'' the annotation.

\paragraph{Modelling.}
Table~\ref{tab:exp1} also reports the prediction performance on
appraisal classes using RoBERTa.  We observe that the performances on
\corpusVis are higher than on \corpusHidden ($+$.03pp on micro \F, .01
on macro \F). This is in line with our assumptions, but the
improvement is actually lower than we expected, given the more
substantial difference in inter-annotator agreement. However, for
\certainty ($+$.09) and \effort ($+$.18), the change is
substantial. \textit{Attentional Activity}, which shows a high
increase in agreement, has a small decrease in modelling performance
($-$.04). \textit{Control}, which does not improve in agreement, has a
considerable loss in prediction performance ($-$.15).

From this experiment, we conclude that the annotation is more reliable
with access to the emotion: this reflects on the modelling results,
and it does so to different extents for different emotions.

\begin{table*}[t]
  \centering\small
  \setlength{\tabcolsep}{7.4pt}
\begin{tabular}{lccc ccc ccc ccc}
  \toprule
  & \multicolumn{3}{c}{deISEAR} & \multicolumn{3}{c}{ISEAR} & \multicolumn{3}{c}{TEC}&
  \multicolumn{3}{c}{Blogs}\\
  \cmidrule(r){2-4}\cmidrule(rl){5-7}\cmidrule(rl){8-10}\cmidrule(l){11-13}
  Appraisal  & P & R& \F & P & R & \F & P & R & \F & P & R & \F \\
  \cmidrule(r){1-1}\cmidrule(r){2-4}\cmidrule(rl){5-7}\cmidrule(rl){8-10}\cmidrule(l){11-13}
  Attentional Activity   & 79 &68 &73 & 83 &82 &83 & 90 &91 &91 & 94 &94 &94\\
  Certainty              & 79 &90 &84 & 89 &92 &90 & 87 &89 &88 & 97 &97 &97\\
  Anticipated Effort     & 88 &93 &91 & 94 &95 &94 & 74 &68 &71 & 91 &93 &92\\
  Pleasantness           & 80 &69 &77 & 91 &90 &91 & 85 &86 &86 & 97 &97 &97\\
  Responsibility/Control & 80 &69 &74 & 88 &85 &86 & 79 &79 &79 & 94 &96 &95\\
  Situational Control    & 73 &69 &71 & 83 &81 &82 & 79 &79 &79 & 88 &86 &87\\
  \cmidrule(r){1-1}\cmidrule(r){2-4}\cmidrule(rl){5-7}\cmidrule(rl){8-10}\cmidrule(l){11-13}
  Macro \avg             & 81 &76 &78 & 88 &87 &88 & 82 &82 &82 & 94 &94 &94\\
  Micro \avg             & 82 &81 &81 & 89 &89 &89 & 84 &84 &84 & 94 &95 &94\\
  \toprule
\end{tabular}
\caption{Experiment 2, Generalization to other corpora. All results
  are averages across 3$\times$10 cross validations. Note that the last three
  columns from Table~\ref{tab:exp2} correspond to the same setting as
  it is in here.}
\label{tab:exp3}
\end{table*}

\subsection{Experiment 2: Automatic Annotation}
\paragraph{Inter-Annotator Agreement.}
We now evaluate the rule-based annotation procedure,
in which appraisal classes are purely assigned by the automatic
procedure, shown in Table~\ref{tab:automatic_annotation}.

The agreement between the rule-based annotation \corpusAuto and both
manual annotations is shown in Table~\ref{tab:exp2}. As
expected, we observe a higher agreement with \corpusVis.
Again, the
differences are not equally distributed across emotions and they
resemble the changes in the other experiments, but agreement
is lower between \corpusAuto and the manual annotations than between
the latters, suggesting that the automatic process does not lead to
the same conceptual annotation\footnote{These
labels could be compared to humans' only on \corpusVis, which
turned out more reliable. We also consider \corpusHidden because it 
represents standard emotion annotation procedures, 
where judges assess texts without further information.}.

\paragraph{Modelling.}
To answer the question how well one model trained on rule-based
annotations performs on manual annotations, we test the model three
times: on \corpusVis, \corpusHidden, and \corpusAuto.
The right side of Table~\ref{tab:exp2} reports the results. Note that
\responsibility and \control have been merged, as explained in the
experimental setting. 

We see that the highest macro average \F is non-surprisingly achieved
when testing on \corpusAuto (.90\F). When testing the same model on
\corpusVis, the performance drops by 12pp (.78\F), but is still
substantially higher than for the corpus in which the emotions were
not available to the annotators (.72\F). Note that the performance of
.78\F (\corpusVis) and .74\F (\corpusHidden) are not too different
from the model trained on manually annotated data, with .80\F and
.79\F. We therefore conclude that automatically labeling a corpus with
appraisal dimensions leads to a meaningful model.

\paragraph{Corpus Generalization.}
Finally, we apply the automatic labeling procedure
to other emotion corpora. Results are in
Table~\ref{tab:exp3}. 

Given the different nature of the domains and
languages (German vs.\ English -- deISEAR/enISEAR; tweets vs.\
blog texts -- TEC vs. Blogs), these numbers cannot be directly
compared, but we can observe that they are comparably high, similar
to the other experiments. We carefully infer (without having compared
the prediction on these corpora to manual annotations) that this is an
indicator that automatic annotation of appraisal dimensions also works
across different corpora and languages.

\subsection{Model Performance Notes and Comparison to Original Data Annotation}
The data that we use was made available to support
appraisal-based research in emotion analysis.
It consists of the same instances
we annotated in \newcite{Hofmann2020}. However, in this previous
work, each instance was judged by three annotators,
who did not have access to the emotion labels of the texts, and the experiments 
 have been performed on labels obtained with the
majority vote of the annotators.
Instead, for the current experiments, the labels by only one annotator on
all instances have been used. Therefore, the experiments of the two
papers are not strictly comparable. In addition, \newcite{Hofmann2020}
adopted a CNN-based classifier. Brief, there are two sources for
non-comparability in our experiments: different
label sets and different models. We aimed at leveraging
 a more state-of-the-art transformer-based model, but at the same time, we needed
different label sets to better understand the appraisal annotation processes.

For transparency reasons, we show the performance of our RoBERTA model
on the original labels against the results by
\newcite{Hofmann2020}. Table~\ref{tab:comparison2} compares the two studies
with respect to appraisals and
Table~\ref{tab:comparison1} with respect to emotion predictions. The emotion
recognition models consist of a text-based model (\taskTE), a
pipeline that first predicts the appraisal and then classifies the
emotion without access to text with a two-layer dense neural network
(\taskTA, \taskAE). To measure the complementarity of these two
settings, a third model is an oracle ensemble
(T\shortarrow{}A\shortarrow{}E + T\shortarrow{}E) which accepts a
prediction as true positive if one of the two models provides the
correct prediction.

On this original data set by \newcite{Hofmann2020}, our model
constitutes a new state of the art. The micro-averaged appraisal
prediction with RoBERTa is 10pp higher than the original CNN-based
model; the emotion classification has similar improvements, and the
overall relation between the model configurations remains comparable.

\begin{table}[t]
  \centering\small
  \begin{tabular}{l ccc ccc ccc}
    \toprule
    & \multicolumn{3}{c}{CNN} & \multicolumn{3}{c}{RoBERTa} \\
    \cmidrule(r){2-4}\cmidrule(l){5-7}
    Appraisal & P & R & \F & P & R & \F \\
    \cmidrule(r){1-1}\cmidrule(r){2-4}\cmidrule(l){5-7}
    Attention      & 81 &84 &82 & 86 &90 &88 \\
    Certainty      & 84 &86 &85 & 87 &94 &91 \\
    Effort         & 68 &68 &68 & 79 &77 &78 \\
    Pleasantness   & 79 &63 &70 & 92 &92 &92 \\
    Responsibility & 74 &68 &71 & 86 &85 &85 \\
    Control        & 63 &49 &55 & 81 &73 &77 \\
    Circumstance   & 65 &58 &61 & 74 &69 &71 \\
    \cmidrule(r){1-1}\cmidrule(r){2-4}\cmidrule(l){5-7}
    Macro \avg & 73 &68 &70 & 83 &83 &83 \\
    Micro \avg & 77 &74 &75 & 84 &85 &85 \\
    \toprule
  \end{tabular}
  \captionof{table}{RoBERTA model performance on predicting appraisals
    on the original data by \newcite{Hofmann2020}, compared to their
    CNN results.}
  \label{tab:comparison2}
\end{table}

\begin{table*}[t]
\centering\small
\begin{tabular}{lccc ccc ccc ccc ccc ccc}
  \toprule
  &\multicolumn{6}{c}{\taskTE}&\multicolumn{6}{c}{\taskTA,\taskAE}&\multicolumn{6}{c}{T\shortarrow{}A\shortarrow{}E + T\shortarrow{}E}\\
  \cmidrule(r){2-7}\cmidrule(rl){8-13}\cmidrule(l){14-19}
  & \multicolumn{3}{c}{CNN} & \multicolumn{3}{c}{RoBERTa}
  & \multicolumn{3}{c}{CNN}& \multicolumn{3}{c}{RoBERTa}&
  \multicolumn{3}{c}{CNN}&\multicolumn{3}{c}{RoBERTa}\\
  \cmidrule(r){2-4}\cmidrule(rl){5-7}\cmidrule(rl){8-10}\cmidrule(rl){11-13}\cmidrule(rl){14-16}\cmidrule(l){17-19}
  Emotion  & P & R& \F & P & R & \F & P & R & \F & P & R & \F & P & R & \F  & P & R & \F \\
  \cmidrule(r){1-1}\cmidrule(r){2-4}\cmidrule(rl){5-7}\cmidrule(rl){8-10}\cmidrule(rl){11-13}\cmidrule(rl){14-16}\cmidrule(l){17-19}
  Anger    & 51 &52 &52 & 62 &62 &62 & 34 &62 &44 & 44 &72 &54 & 66&81&73 & 72& 85& 78 \\
  Disgust  & 65 &63 &64 & 70 &71 &71 & 59 &34 &43 & 65 &38 &48 & 78&68&73 & 86& 74& 80 \\
  Fear     & 69 &71 &70 & 80 &82 &81 & 55 &55 &55 & 65 &76 &70 & 76&77&77 & 85& 92& 88 \\
  Guilt    & 47 &42 &44 & 60 &60 &60 & 38 &50 &43 & 50 &68 &58 & 60&63&62 & 72& 80& 76 \\
  Joy      & 74 &80 &77 & 92 &96 &94 & 77 &69 &72 & 92 &95 &93 & 79&80&80 & 94& 98& 96 \\
  Sadness  & 69 &67 &68 & 82 &81 &81 & 58 &40 &47 & 74 &55 &63 & 74&70&72 & 87& 84& 86 \\
  Shame    & 44 &45 &45 & 57 &54 &55 & 36 &24 &29 & 51 &23 &32 & 58&51&54 & 77& 59& 67 \\
  \cmidrule(r){1-1}\cmidrule(r){2-4}\cmidrule(rl){5-7}\cmidrule(rl){8-10}\cmidrule(rl){11-13}\cmidrule(rl){14-16}\cmidrule(l){17-19}
  Macro \avg&60 &60 &60 &72 &72 &72 & 51 &48 &48 & 63 &61 &60 & 70&70&70 & 82& 82& 82 \\
  Micro \avg&   &   &60 &   &   &72 &    &   &48 &    &   &61 &   &  &70 &   &   & 82\\
  \toprule
\end{tabular}
\caption{Comparison of the CNN \cite{Hofmann2020} and our RoBERTa
  model on the Text-to-Emotion baseline (\taskTE), the pipeline
  experiment (\taskTA,\taskAE) and the oracle ensemble experiment
  (T\shortarrow{}A\shortarrow{}E + T\shortarrow{}E). These experiments
  follow the model configurations by \newcite{Hofmann2020}.}
\label{tab:comparison1}
\end{table*}

\begin{table*}[t]
  \centering\small
  \begin{tabularx}{\linewidth}{lcllX}
    \toprule
    ID & Score & Emotion & Aggr.\ Change & Input Text\\
    \cmidrule(r){1-1}\cmidrule(rl){2-2}\cmidrule(rl){3-3}\cmidrule(rl){4-4}\cmidrule(l){5-5}
    1  & $+$4 & Fear & $+$e $+$p $+$r $+$ci &
                                  I felt \ldots\ when I was abseiling down a
                                  cliff-face.
    \\
    2  & $+$3 & Joy & $+$ce $+$r $+$ci &
                             I felt \ldots\ when I got a new job.
    \\
    3  & $\pm$0 & Guilt & &
                       I felt ... when I participated in gossip at
                       work.
    \\
    4  & $-$2 & Shame & $-$a $-$ce $-$r $+$ci &
                                  I felt \ldots\ when I found out that
                                  my daughter had been having a difficult time
                                  and I didn't realise straight away what she
                                  was going through.
    \\
    5  & $-$2 & Anger & $-$a $-$e &
                              I felt \ldots\ when we were charged by a care
                              home for the three months after my father had
                              died, even though we had emptied his room the
                              day after his death.
    \\
    6  & $-$3 & Fear & $-$a $-$ce $-$r &
                                 I felt \ldots\ when cycling home after
                                 a long ride one evening, unaware how dark it
                                 had become, and thus relying on some very
                                 weak led lights that I'd never tested in
                                 complete darkness - I could barely see ten
                                 feet ahead of me.
    \\
 \bottomrule
  \end{tabularx}
  \caption{Examples of differences between annotations with masked and
    visible emotion labels. $+$ and $-$ indicate the agreement and the
    disagreement on a specific dimension which is reached after making
    the emotion visible. The score is the sum of agreement changes,
    either improvements ($+1$ for each dimension) or degradation
    ($-1$). a: \attention, ce: \certainty, e: \effort, p:
    \pleasantness, r: \responsibility, ci: \circumstance.}
  \label{tab:qualitative_analysis}
\end{table*}

\section{Qualitative Analysis}
To better understand how revealing emotions affects the annotations in
Experiment 1, we provide some concrete examples. Table
\ref{tab:qualitative_analysis} reports instances from enISEAR. We show
for \textit{which appraisal variables the agreement changes}, by
marking the appraisal with $+$ or $-$. For instance,
$-$\attention means that the annotators came to disagree on
that appraisal dimension when the emotion was uncovered, while
$+$\pleasantness indicates that they came to agree thanks to the
knowledge of the emotion label. Note that $+$ does not mean that the
dimension was marked as 1 by both annotators.
The examples are sorted by the sum of changes in agreement.

In Example (1) an event is described in a way which leaves open if
there is \responsibility, \pleasantness, \effort, and even if the
experiencer is entirely certain about what is happening. With the
knowledge that the emotion is fear, it becomes clear that the
situation does involve \effort, is not \textit{pleasant}, and that
\circumstance is not likely. The annotators also agree here that there
is \responsibility involved, which is likely an interpretation based
on world-knowledge.

In the situation of getting a new job (Example (2)), knowing the
emotion adds agreement regarding certainty, which is in line with
added agreement that the person was responsible.
Example (3) is an instance in which the situation was already
entirely clear without knowing the emotion -- the experiencer
participated in gossip. That is a certain, non-pleasurable situation
(when recognizing this) which is under their own control. Knowledge
that guilt has been experienced does not add anything.

Example (4) shows that complex situations cause more disagreement in
annotations which are not necessarily resolved by knowing the
emotion. The described event is about a negative emotion felt because
the experiencer did not recognize the bad mood of the daughter.
Annotators come to disagree about \attention, \certainty and
\responsibility.

Example (5) describes another situation in which a negative
event is discussed. Knowledge of the emotion puts a clear focus away
from the sad part of the description (father dying) and puts it on
something that causes anger. However, this shift does not resolve
appraisal disagreements but indeed adds on top of them, with \attention
and \effort.

Finally, Example (6) is another long description with annotators'
focus on different aspects, one on the darkness (hence, no
\responsibility), the other on the cycling
(\responsibility). Here, knowledge of the emotion does not change
the interpretability of the event. However, it informs on
 which part of the described situation the original author
focused on.

These examples show that \corpusVis helps solve ambiguities when
events can be associated to multiple emotions, other times it helps
people give more weight to specific portions of texts. In the first
case agreement tends to be reached more easily.

\section{Conclusion \& Future Work}
We analyzed how to build corpora of text annotated with appraisal
variables and we evaluated how well such concepts can be modelled.  By
doing so, we brought together emotion analysis and a strand of emotion
research in psychology which has received little attention from the
computational linguistics community. We propose that in addition to
well-established approaches to emotion analysis, like affect-oriented
dimensional approaches or classification into predefined emotion
inventories, psychological models of appraisals will be considered in
future emotion-based studies, particularly those relying on
event-oriented resources.

The use of appraisals is interesting from a \emph{theoretical perspective}:
motivated by psychology, we leveraged the cognitive mechanisms underlying
emotions, thus accounting for many complex patterns in which humans
appraise an event and emotionally react to it. In this light,
it is interesting in itself that our annotators were able to
empathically reconstruct the event appraisals
experienced by others, even without knowing their emotion.

From a \emph{practical perspective}, appraisal annotations are less
prone to being poorly chosen for particular domains, in comparison
to regular emotion classes, as the actual
feeling develops based on the cognitive evaluation of an event. We
also have shown that event descriptions alone might not be sufficient
to properly annotate the hypothetical appraisal of the experiencer
(which is, however, also an issue with traditional emotion analysis
annotations and models -- we cannot look into the feeler, we deal with
private states). This shows and is presumably the reason that
additional context (e.g., the emotion label) is required.

Some implications for future research and developments follow: ideally,
appraisal (and emotion) annotations should stem directly from the
experiencer. This is not doable in many NLP settings. For instance,
when analyzing literature, it is impossible to ask fictional
characters for their current event appraisal. However, we presume that
settings on social media might be realistic, for instance by probing
 appropriate distant labeling methods, e.g., a careful
choice of hashtags.  If this is unfeasible, because text
authors do not disclose their appraisals, the available emotion labels
still represent a valuable source of information: as we have shown, they
can guide
the interpretation of the described events, and hence, the way in which
these are post-assigned appraisal dimensions.

Finally, we provided evidence that even if a model is trained on
automatically obtained appraisal labels, it is still capable of
substantial performance.  Therefore, we conclude that more corpora
with appraisal dimensions from different languages and domains should
be created from scratch. In the meantime, one can build on top of the
rich set of available emotion corpora and automatically create
appraisal-annotated resources out of them.

\section*{Acknowledgements}
This work was supported by Deutsche Forschungsgemeinschaft (projects
SEAT, KL 2869/1-1 and CEAT, KL 2869/1-2) and has partially been
conducted within the Leibniz WissenschaftsCampus T\"ubingen
``Cognitive Interfaces''.  We thank Kai Sassenberg and Laura
Oberl\"ander for inspiration and fruitful discussions.

\bibliographystyle{acl_natbib}
\bibliography{lit}

\pagebreak

\appendix


\section{Corpus Statistics for Manual Annotations of enISEAR}
Table~\ref{tab:manual_annotation_counts} shows the corpus statistics
for the manually annotated corpora. In particular, it depicts how the
availability of the emotion to the annotators influences the
distribution of appraisal labels. The same numbers are also shown in a
comparative manner in Figure~\ref{fig:dist}. For most appraisal
dimensions, the annotations are more specific, narrower, across the
counts for emotions and mostly manifest in a lower number of those. An
exception is \certainty, which shows higher counts for all emotions, and
\effort, which receives higher counts for shame, sadness, fear, and
guilt, but not for anger.\\[1cm]

\newcommand{\mco}[1]{\multicolumn{1}{c}{\rt{#1}}}
\begingroup
\centering\footnotesize
 \setlength{\tabcolsep}{5pt}
\begin{tabular}{r|rrrrrrrr}
  \toprule
  \multicolumn{1}{c}{}&& \multicolumn{7}{c}{Appraisal Dimension}\\
  \cmidrule(l){3-9}
  \multicolumn{1}{c}{}& Emotion  & \mco{Attention} & \mco{Certainty} & \mco{Effort}
  & \mco{Pleasant} & \mco{Respons.} & \mco{Control}
  & \mco{Circum.} \\
  \cmidrule(lr){2-2}\cmidrule(rl){3-3}\cmidrule(rl){4-4}\cmidrule(rl){5-5}
  \cmidrule(rl){6-6}\cmidrule(rl){7-7}\cmidrule(l){8-8}\cmidrule(l){9-9}
  \multirow{8}{*}{\rt{enISEAR \corpusVis}}
      & Anger    &   141   &   143   &    17   &     0   &     4   &     1   &     3   \\
      & Disgust  &    13   &   143   &    65   &     0   &    14   &     8   &    11   \\
      & Fear     &   126   &    24   &   139   &     0   &    18   &     4   &   115   \\
      & Guilt    &    70   &   141   &   108   &     0   &   141   &    93   &    11   \\
      & Joy      &   143   &   143   &     0   &   143   &    43   &    21   &    18   \\
      & Sadness  &   120   &   141   &   136   &     0   &     4   &     2   &   132   \\
      & Shame    &    10   &   137   &   105   &     0   &   113   &    23   &     4   \\
      \cmidrule(lr){2-2}\cmidrule(rl){3-3}\cmidrule(rl){4-4}\cmidrule(rl){5-5}
      \cmidrule(rl){6-6}\cmidrule(rl){7-7}\cmidrule(l){8-8}\cmidrule(l){9-9}
      & Total    &    623  &    872  &    570  &    143  &    337  &    152  &    294  \\
      \midrule
      \multirow{8}{*}{\rt{enISEAR \corpusHidden}}
      & Anger    &   130   &   113   &    60   &     0   &     8   &     1   &    11 \\
      & Disgust  &    58   &   129   &    35   &     1   &    16   &     7   &    35 \\
      & Fear     &   126   &    13   &   125   &     0   &    33   &    11   &   108 \\
      & Guilt    &    54   &   128   &    29   &     1   &   139   &    85   &    21 \\
      & Joy      &   134   &   125   &     4   &   139   &    55   &    46   &    56 \\
      & Sadness  &   121   &   105   &    69   &     1   &    10   &     3   &   119 \\
      & Shame    &    25   &    98   &    33   &     1   &   113   &    62   &    18 \\
      \cmidrule(lr){2-2}\cmidrule(rl){3-3}\cmidrule(rl){4-4}\cmidrule(rl){5-5}
      \cmidrule(rl){6-6}\cmidrule(rl){7-7}\cmidrule(l){8-8}\cmidrule(l){9-9}
      & Total    &    648  &    711  &    355  &    143  &    374  &    215  &    368 \\
      \toprule
\end{tabular}
\captionof{table}{\label{tab:manual_annotation_counts} Manual
  Annotation: Instance counts across emotions and appraisal
  annotations.}
\endgroup

\end{document}